\documentclass{article}
\usepackage{graphicx} 
\usepackage{multicol}
\usepackage{indentfirst} 
\usepackage[margin=0.8in]{geometry}
\usepackage{amsmath}
\usepackage{caption}
\usepackage[colorlinks,linkcolor=blue]{hyperref}

\title{\textbf{Multimodal Fusion with Pre-Trained Model Features in Affective Behaviour Analysis In-the-wild} }
\date{} %

\author{
  \begin{tabular}{c}
    Zhuofan Wen$^{1,2\ast}$, Fengyu Zhang$^{1,2\ast}$, Siyuan Zhang$^{1,2}$, Haiyang Sun$^{1,2}$, Mingyu Xu$^{1,2}$, \\
    Licai Sun$^{1,2}$, Zheng Lian$^{2}$, Bin Liu$^{1,2\dagger}$, Jianhua Tao$^{3,4}$ \\
    \multicolumn{1}{p{\textwidth}}{\centering\normalsize
    $^1$University of Chinese Academy of Sciences, China \\
    $^2$The State Key Laboratory of Multimodal Artificial Intelligence Systems, Institute of Automation, Chinese Academy of Sciences \\
    $^3$Department of Automation, Tsinghua University, China \\
    $^4$Beijing National Research Center for Information Science and Technology, Tsinghua University
    \\
    \vspace{0.5em}
    $^\ast$These authors are equal contributors to this work. \\
    $^\dagger$Corresponding author
    }
  \end{tabular}
}

\begin{document}

\maketitle
\begin{multicols*}{2}

\section*{\centerline{Abstract}}
Multimodal fusion is a significant method for most multimodal task, With the recent surge in the number of large pre-trained models, combine both multimodal fusion methods and pre-trained model features can archive outstanding performance in many multimodal task. In this paper, we present our approach which leveraging both advantages for addressing the task of Expression (Expr) Recognition and Valence-Arousal (VA) Estimation. We evaluate the Aff-Wild2 database using pre-trained models, then extract the final hidden layers of the models as features. Following preprocessing and interpolation or convolution to align the extracted features, different models are employed for modal fusion. Our code are avalible on \href{code}{https://github.com/FulgenceWen/ABAW6th}.
\section{Introduction}
\begin{figure*}
    \centering
    \includegraphics[width=1\linewidth]{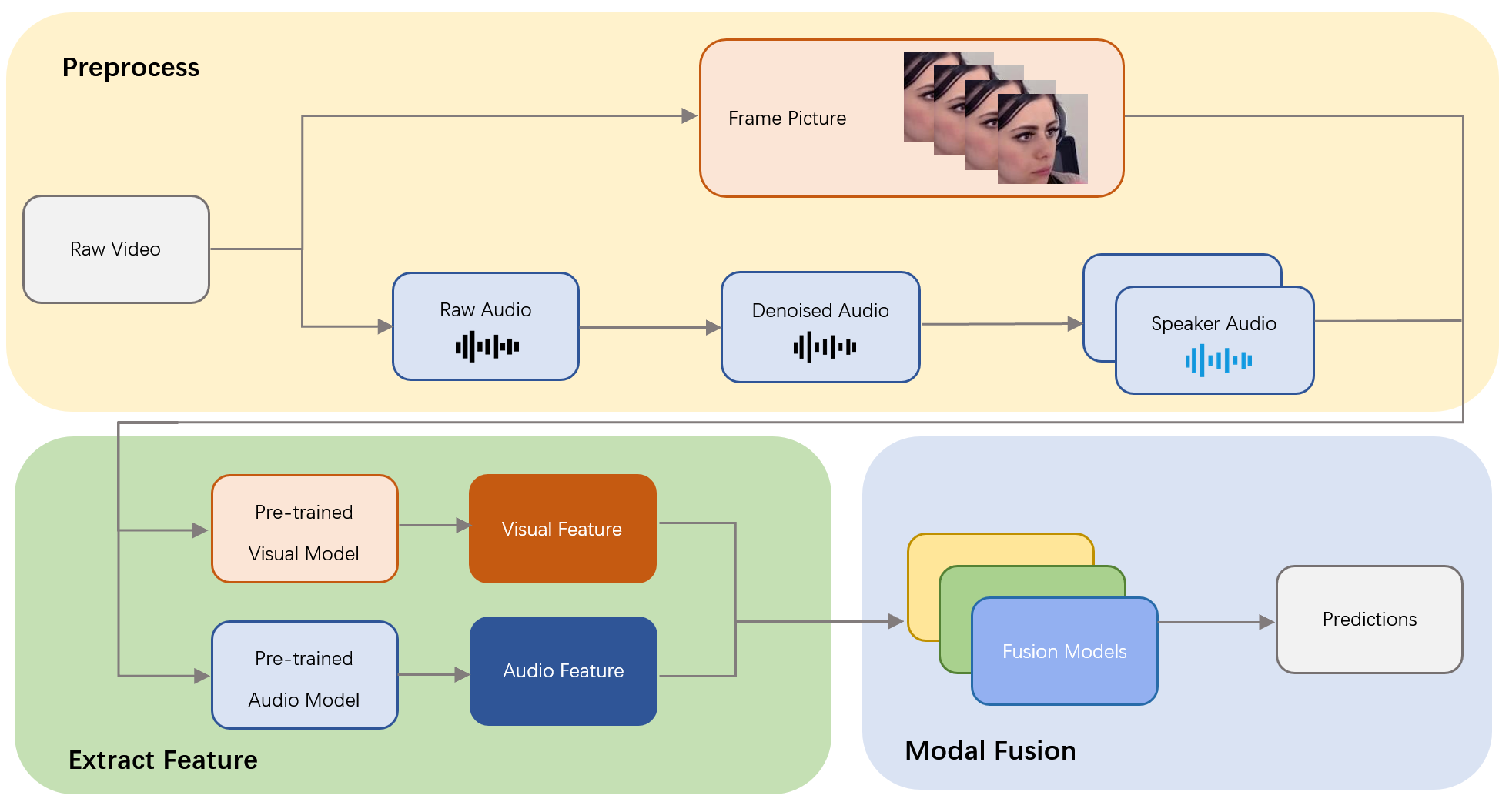}
    \caption{The block diagram illustrates our proposed method, detailing the primary steps involved in data processing, feature extracting and modal feature fusion.}
    \label{fig:enter-label}
\end{figure*}
In recent years, there has been a surge in the introduction of pretrained models. These models are meticulously designed and exceptionally robust, enabling us to address multiple tasks seamlessly with a single model. Consequently, downstream tasks no longer necessitate the redesigning of models or specialized efforts. \cite{kollias2023abaw} \cite{kollias2022abaw}

In this paper, we present our solutions to the 6th Affective Behavior Analysis in the Wild (ABAW) competition.\cite{kollias2023abaw2}\cite{kollias20246th} As Figure.1 shows, we first 
use different strategy preprocess raw data, than explore a series of features from the audio, visual, and text modality. For the VA challenge, we utilize the multimodal cyclic Translation Network (MCTN), Memory Fusion Network (MFN) and a simple attention network\cite{lian2023mer} to capture complex temporal dynamics in the feature sequence. For the Expr challenge\cite{kollias2023multi} , we have chosen Mobilenetv3 as the foundational component of our network, leveraging its capabilities as a powerful backbone. To extract strong and reliable representations from this backbone, we incorporate a transformer encoder as the embedding layer. Our approach involves utilizing multi-head attention, where we expand the space of the attention heads and allow for flexible embedding dimensions.\cite{phan2022facial} By employing this architecture, our model achieves superior performance compared to the baseline on the validation set. To boost the model’s performance, we incorporate tactics such as pseudo-labeling, label smoothing, and other techniques to further enhance performance.\cite{kollias2021analysing}

\section{VA Estimation Method}
The Valence-Arousal (VA) Estimation challenge requires participants to predict emotional dimensions (i.e., arousal and valence) in a time-continuous manner (i.e., every 0.25s).\cite{kollias2021affect}
Our proposed method mainly consists of three modules: preprocess and feature extraction module, loss function, and fusion module. The preprocess module denoise and separate different speaker information and the feature extraction module extracts a variety of features from the audio, visual, and text modality, after that different modality data need to align as the same length for sequence task. Then, we utilize various fusion models to integrate multimodal information.\cite{kollias2021distribution}
To guide the model training, the concordance correlation coefficient (CCC) loss is employed for Valence-Arousal (VA) Estimation. Finally, we incorporate tactics such as pseudo-labeling, label smoothing, and other techniques to further enhance performance.
\subsection{Preprocess and Feature Extraction}
The 6th Affective Behavior Analysis in the Wild (ABAW) competition provide Aff-Wild2 database. \cite{kollias2019deep} The videos in this dataset encompass multi-person conversations, speeches, or mixed videos featuring short clips along with audiences, adding numerous challenges for the upcoming task. \cite{kollias2020analysing}

To handle this complex and diverse data, which includes multi-person conversations, speeches, and videos with mixed content and various noises, We adopt difference strategies on different modality. 

For visual, we utilize various visual pre-trained models to extract cropped-aligned face frame features(i.e., Clip-large, Resnet, Senet, Eva02-large.). This is because face frames generally contain fewer noises and are easier to align with features from other modalities. We then evaluate the performance of these features using uni-modal models and select the best-performing ones.

For audio, we employ denoising techniques for the audio and separate speakers within the audio. Subsequently, we utilize different pre-trained audio models to extract audio features(i.e., Wavlm, Whisperv2, Whisperv3). Once the audio is clear, we employ interpolation or convolution methods to align the audio features with the visual features.

For text, Based on our experiments and the results from previous ABAW competitions, we observed that the performance improvement of the text modality was not significant. Therefore, we did not proceed with further fusion of text modality data.\cite{kollias2019expression}
\subsection{Loss Function}
As suggested by the last year’s winner in ABAW 5th Valence-Arousal Estimation challenge, the concordance correlation coefficient (CCC) loss generally outperforms either mean squared loss or L1 loss; therefore, we employ the CCC loss as the loss function for this sub-challenge. It is defined as follows:
\begin{equation}
    L_{CCC} = 1 - CCC
\end{equation}
\begin{equation}
    CCC = \frac{2\rho\sigma_{\hat{Y}}\sigma_Y}{\sigma_{\hat{Y}}^2 + \sigma_Y^2 + (\mu_{\hat{Y}} - \mu_Y)^2} 
\end{equation}
where $\mu_{\hat{Y}}$ and $\mu_Y$ are the mean of the prediction $\hat{Y}$ and the label $Y$, respectively. $\sigma_{\hat{Y}}$ and $\sigma_Y$ are their standard deviations. $\rho$ is the Pearson correlation coefficient (PCC) between $\hat{Y}$ and $Y$.

\subsection{Multimodal Fusion}
For the VA challenge, we utilize the multimodal cyclic Translation Network (MCTN), Memory Fusion Network (MFN) and a simple attention network to assess the performance of feature and model combinations. Additionally, we incorporate tactics such as pseudo-labeling, label smoothing, and other techniques to further enhance performance.

The Multimodal Cyclic Translation Network (MCTN) is a neural model designed to learn robust joint representations through modality translations. Illustrated in Figure 2, MCTN provides a comprehensive overview for two modalities. This model leverages the fundamental concept that translating from a source modality \( X_S \) to a target modality \( X_T \) generates an intermediate representation capturing joint information between the two modalities. 

The Memory Fusion Network (MFN) is a recurrent model for multi-view sequential learning that comprises three main components: 
\begin{itemize}
    \item A system of LSTMs: Multiple LSTM networks, each encoding view-specific dynamics.
    \item Delta-memory Attention Network: Specialized attention mechanism for discovering cross-view interactions.
    \item Multi-view Gated Memory: Memory unit for storing cross-view interactions over time. Figure 1 outlines the MFN pipeline and components.
\end{itemize}

The input to MFN is a multi-view sequence with the set of \( N \) views, each of length \( T \). For example, sequences can consist of language, video, and audio for \( N = \{l, v, a\} \). The input data of the \( n \)th view is denoted as: \( x_n = [x^{t}_{n} : t \leq T, x^{t}_{n} \in {R}^{d_{xn}}]\) is the input dimensionality of the $n$th view input $x_n$.

\section{Expr Recognition Method}
The task of Expression (Expr) Recognition is to recognize 6 basic expressions (i.e., anger, disgust, fear, happiness, sadness, surprise), plus the neutral state, plus a category 'other' that denotes expressions/affective states other than the 6 basic ones. \cite{kollias2019face} 
We utilize a transformer encoder with multi-head attention as the embedded layer to generate sequence representations. The transformer encoder plays a crucial role in encoding robust representations for the backbone of our model. Additionally, we employ the pre-trained model (i.e, Mobilenetv3, Efficientnetv2) as the backbone for our proposed network. The input images have a size of 112x112x3, and we extract a backbone with 1024 features using the flattened layer of the pre-trained model. This backbone is reshaped to (batch size, sequence, feature) and fed into the transformer encoder. For this study, we set the sequence length to 64.

\subsection{Residual Network }

The Residual Network (ResNet) is a deep neural network model designed to address the problem of vanishing gradients during training. ResNet achieves this by incorporating residual connections, enabling the network to learn across layers more effectively.\cite{he2016deep}

In ResNet, each building block consists of two convolutional layers and a residual connection. The residual connection adds the input feature mapping directly to the output feature mapping of the building block, allowing the network to learn residual mappings (i.e., the differences between input and output). The residual block is formulated as follows:
\begin{equation}
    y=\mathcal{F}(x,W)+x
\end{equation}

where $x$ represents the input feature vector,
 $\mathcal{F}$ is the mapping function of the building block,
 $W$ represents the weights of the building block. 

Compared to conventional neural networks, ResNet's key improvement lies in the inclusion of skip connections, which allow information to flow directly from earlier layers to later layers, effectively alleviating the problem of vanishing gradients.

To further increase the depth of the network, ResNet stacks multiple residual blocks together, forming a deep architecture.

In this work, we set $\mathcal{F}$ as a large-scale Transformer encoder structure and $x$ as the representation of the MobileNetV3 backbone. $\mathcal{F}(x)$ represents the residual learned by the network block between the original network block and the input feature $x$.

\subsection{Transformer Encoder}

We utilize the Transformer Encoder to learn temporal features. The Encoder part of the Transformer consists of six identical encoder layers. Taking a single layer as an example, each encoder has two sub-layers: the Multi-Head Attention layer and the Position-wise Feed-Forward Network. After each sub-layer, there are residual connections and layer normalization operations. The Transformer Encoder maps the input sequence to a representation feature. \cite{vaswani2017attention} In this study, we employ the Transformer Encoder as an embedding layer to extract robust representation features from the MobileNetV3 backbone.

\subsection{Loss Function}
We employ F1-loss as the loss function. Empirically, we find that F1-loss usually performs better than the standard cross-entropy loss. The F1 loss is calculated as follows:
\begin{equation}
    L_{f1} = 1 - f1 
\end{equation}
\begin{equation}
    TP = \sum_{i=1}^{N} \hat{Y}_i \odot Y_i
\end{equation}
\begin{equation}
    FP = \sum_{i=1}^{N} (1 - \hat{Y}_i) \odot Y_i 
\end{equation}
\begin{equation}
    FN = \sum_{i=1}^{N} \hat{Y}_i \odot (1 - Y_i) 
\end{equation}
\begin{equation}
    F1 = \frac{2TP}{2TP + FP + FN} 
\end{equation}
\begin{equation}
    f1 = \mu_{F1} 
\end{equation}
where $\hat{Y}_i$ and $Y_i$ are the prediction and label for the $i$th segment, $\odot$ denotes the Hadamard product, $N$ is the number of segments in one training batch, $\mu_{F1}$ is the mean of F1 over five discrete sentiment classes.

\section{Experiment}
Table 1 presents the experimental results of our proposed method on the validation set of the VA and Table 2 presents the experimental results of Expr Challenge. We utilize the Concordance Correlation Coefficient (CCC) as the evaluation metric for both valence and arousal prediction, while the F1-score is employed to evaluate the result of the Expr challenge. 

For the VA Challenge, an augmented version of the Aff-Wild2 database will be utilized.\cite{zafeiriou2017aff} This database is audiovisual (A/V), in-the-wild, and comprises a total of 594 videos containing approximately 3 million frames, with annotations for valence and arousal across 584 subjects. For this challenge, the dataset is split into 356 videos for training, 76 videos for validation, and 162 videos for testing.
The performance measure ($P$) is the mean Concordance Correlation Coefficient (CCC) of valence and arousal:

\[
P = \frac{{CCC_{arousal} + CCC_{valence}}}{2}
\]

For Expr Challenge, the Aff-Wild2 database is audiovisual (A/V), in-the-wild and in total consists of 548 videos of around 2.7M frames that are annotated in terms of the 6 basic expressions (i.e., anger, disgust, fear, happiness, sadness, surprise), plus the neutral state, plus a category 'other' that denotes expressions/affective states other than the 6 basic ones. For this challenge, the dataset comprises 246 videos for training, 70 videos for validation, and 232 videos for testing.
The performance measure ($P$) is the average F1 Score across all 8 categories:

\[ P = \frac{\sum F1}{8} \]

As depicted in the table, our proposed method significantly outperforms the baseline. These results underscore the effectiveness of our approach, which leverages multimodal fusion methods and pre-trained model features to capture complex temporal dynamics in the feature sequence, thereby enhancing accuracy.

\begin{center}
    \begin{tabular}{ cccc }
        \hline
        \textbf{Model}& \textbf{Visual Features} & \textbf{Audio Features} & \textbf{CCC} \\
        \hline
        Attention & Resent&None & 0.5342 \\
        Attention & Senet&None & 0.5513 \\
        Attention & CLip-large&None & 0.6347 \\
        Attention & Eva02&None & 0.6082 \\
        Attention & CLip-large&Wavlm & 0.6523 \\
        MFN & CLip-large &Wavlm & 0.6874 \\
        MFN & Eva02&Wavlm & 0.6523 \\
        MCTN & CLip-large &Wavlm & 0.6943 \\
        MCTN & Eva02&Wavlm & 0.6645 \\
        \hline
    \end{tabular}
    \captionof{table}{Model and Features Combination with Val Scores on VA challenge. parameters: hidden dim: 256, dropout: 0.2, loss weight: 0.5, grad clip: 0.8, lr: 1e-3}\label{tab:model_features}
\end{center}

\begin{center}
    \begin{tabular}{ cccc }
        \hline
        \textbf{Model}& \textbf{Visual Features} & \textbf{lr} & \textbf{F1} \\
        \hline
        Attention &efficientnet v2 &0.001& 0.255 \\
        Attention &efficientnet v2 &0.0005& 0.274 \\
        Attention &mobilenet v3 &0.001& 0.289 \\
        Attention &mobilenet v3 &0.0005&0.263 \\
        \hline
    \end{tabular}
    \captionof{table}{Model and Features Combination with Val Scores on Expr challenge. Parameters: batchsize=16, lr: 1e-3, epoch=30}\label{tab:model_features}
\end{center}

\section{Conclusion}
In this paper, we outline our approach to the VA challenge and Expr challenge of the 6th ABAW. We explore various pre-trained features from three common modalities: audio, visual, and text. For the fusion model, we employ Attention, MFN, and MCTN models to integrate modality information. For the Expr challenge, we utilize MobileNetV3 to extract visual features, employ the Transformer Encoder to learn temporal features, and connect them using the Residual Network.The results of our best submissions demonstrate that our proposed method outperforms the baseline system in both the VA challenge and Expr challenge. Future work will involve experimenting with more advanced fusion methods and alignment techniques.
\bibliographystyle{unsrt}
\bibliography{references}
\end{multicols*}
\end{document}